\documentclass[12pt, letterpaper]{article}

\usepackage[margin=1in]{geometry}
\usepackage{amsmath, amssymb, amsfonts}
\usepackage{graphicx}
\usepackage{booktabs}
\usepackage{array}
\usepackage{tabularx}
\usepackage{multirow}
\usepackage{xcolor}
\usepackage{hyperref}
\usepackage{url}
\usepackage{enumitem}
\usepackage{caption}
\usepackage{subcaption}
\usepackage{float}
\usepackage{microtype}
\usepackage{parskip}
\usepackage[numbers,sort&compress]{natbib}
\usepackage{pifont}
\usepackage{tikz}
\usetikzlibrary{shapes.geometric, arrows.meta, positioning, fit, backgrounds, calc}
\usepackage{pgfplots}
\pgfplotsset{compat=1.18}
\usepackage{listings}
\usepackage{algorithm}
\usepackage{algpseudocode}
\usepackage{mdframed}
\usepackage{tcolorbox}
\tcbuselibrary{skins, breakable, listings}
\usepackage{authblk}

\usepackage{caption}
\usepackage{float}
\floatstyle{plain}                   
\newfloat{codeblock}{H}{loc}
\floatname{codeblock}{Example}

\hypersetup{
  colorlinks=true,
  linkcolor=blue!60!black,
  citecolor=red!70!black,
  urlcolor=blue!70!black,
}

\definecolor{vtMaroon}    {HTML}{861F41}   
\definecolor{vtOrange}    {HTML}{E5751F}   
\definecolor{vtStone}     {HTML}{75787B}   
\definecolor{vtTeal}      {HTML}{508590}   
\definecolor{vtPurple}    {HTML}{642667}   
\definecolor{vtGrey}      {HTML}{D7D2CB}   
\definecolor{vtSmoke}     {HTML}{E5E1E6}   

\definecolor{demlat}      {HTML}{861F41}   
\definecolor{vision}      {HTML}{508590}   
\definecolor{reservoir}   {HTML}{E5751F}   
\definecolor{analysis}    {HTML}{642667}   
\definecolor{optimize}    {HTML}{75787B}   

\definecolor{lightdemlat}    {HTML}{F4E8EC}
\definecolor{lightvision}    {HTML}{E8F2F4}
\definecolor{lightreservoir} {HTML}{FDF0E4}
\definecolor{lightanalysis}  {HTML}{EDE6EE}
\definecolor{lightoptimize}  {HTML}{EDEDEE}
\definecolor{codebg}         {HTML}{F8F8FA}

\usepackage{listings}
\usepackage{xcolor}

\definecolor{lstbg}    {HTML}{F8F8F8}
\definecolor{lstkw}    {HTML}{007020}   
\definecolor{lststr}   {HTML}{4070A0}   
\definecolor{lstcom}   {HTML}{60A0B0}   
\definecolor{lstnum}   {HTML}{208050}   
\definecolor{lstbuiltin}{HTML}{007020}  
\definecolor{lstlineno}{HTML}{AAAAAA}   

\tcbset{
  modulebox/.style 2 args={
    enhanced,
    colframe=#1,
    colback=#2,
    colbacktitle=#1,
    coltitle=white,
    fonttitle=\footnotesize\bfseries\ttfamily,
    boxrule=0.6pt,
    arc=3pt,
    left=5pt, right=5pt,
    top=3pt, bottom=3pt,
    toptitle=2pt, bottomtitle=2pt,
    fontupper=\footnotesize,
    before skip=4pt,
    after skip=4pt,
  }
}

\title{\Large\textbf{OpenPRC: A Unified Open-Source Framework for Physics-to-Task Evaluation in Physical Reservoir Computing}}

\author[1,*]{\small Yogesh Phalak}
\author[1,*]{Wen Sin Lor}
\author[1]{Apoorva Khairnar}
\author[2,3]{Benjamin Jantzen}
\author[1]{Noel Naughton}
\author[1]{Suyi Li}

\affil[1]{\footnotesize Department of Mechanical Engineering, Virginia Tech, Blacksburg, VA, USA}
\affil[2]{\footnotesize Department of Philosophy, Virginia Tech, Blacksburg, VA, USA}
\affil[3]{\footnotesize Department of Computer Science, Virginia Tech, Blacksburg, VA, USA}
\affil[*]{\footnotesize To whom correspondence should be addressed; E-mail: yphalak@vt.edu, wensin@vt.edu}

\date{%
  \small \today\\[6pt]
  \small \textbf{Code:} \href{https://github.com/DARE-Lab-VT/OpenPRC-dev}{\texttt{github.com/DARE-Lab-VT/OpenPRC-dev}}
}

\begin{document}
\maketitle
\thispagestyle{empty}

\begin{abstract}
Physical Reservoir Computing (PRC) leverages the intrinsic nonlinear dynamics of physical substrates --- mechanical, optical, spintronic, and beyond --- as fixed computational reservoirs, offering a compelling paradigm for energy-efficient and embodied machine learning. However, the practical workflow for developing and evaluating PRC systems remains fragmented: existing tools typically address only isolated parts of the pipeline, such as substrate-specific simulation, digital reservoir benchmarking, or readout training. What is missing is a unified framework that can represent both high-fidelity simulated trajectories and real experimental measurements through the same data interface, enabling reproducible evaluation, analysis, and physics-aware optimization across substrates and data sources.

We present \textbf{OpenPRC}, an open-source Python framework that fills this gap through a schema-driven physics-to-task pipeline built around five modules: a GPU-accelerated hybrid RK4--PBD physics engine (\texttt{demlat}), a video-based experimental ingestion layer (\texttt{openprc.vision}), a modular learning layer (\texttt{reservoir}), information-theoretic analysis and benchmarking tools (\texttt{analysis}), and physics-aware optimization (\texttt{optimize}). A universal HDF5 schema enforces reproducibility and interoperability, allowing GPU-simulated and experimentally acquired trajectories to enter the same downstream workflow without modification. Demonstrated capabilities include simulations of Origami tessellations video-based trajectory extraction from a physical reservoir, and a common interface for standardized PRC benchmarking, correlation diagnostics, and capacity analysis. The longer-term vision is to serve as a standardizing layer for the PRC community, compatible with external physics engines including PyBullet, PyElastica, and MERLIN.
\end{abstract}


\vspace{0.5em}
\hrule
\vspace{1em}

\section{Introduction}
\label{sec:intro}

Reservoir computing (RC) is a computational paradigm in which a fixed, high-dimensional dynamical system --- the \emph{reservoir} --- maps a time-varying input into a rich feature space from which a simple trained readout extracts the desired output~\cite{Jaeger2001,Maass2002}. The reservoir's internal weights are never updated; only a linear readout is trained, dramatically reducing computational cost relative to fully recurrent networks. This architectural simplicity opens a unique pathway: the reservoir need not be digital at all. Any physical system exhibiting high-dimensional nonlinear dynamics, fading memory, and sufficient separation property can serve as a reservoir~\cite{Tanaka2019,Nakajima2020}.

Physical Reservoir Computing~\cite{Nakajima2020} (PRC) exploits this insight by substituting the software reservoir with a physical substrate whose natural time evolution performs the mapping. Demonstrations span an extraordinary range of materials and scales: nanomagnetic spin textures and domain-wall dynamics~\cite{Allwood2023}, emerging electronic reservoirs~\cite{Liang2024}, active microparticles~\cite{Wang2024}, soft robots and tensegrity structures~\cite{Caluwaerts2013,Nakajima2018}, and bio-inspired underwater propulsors~\cite{Hess2024}. Mechanical systems occupy a particularly attractive corner of this landscape. The nonlinear folding kinematics of origami~\cite{Bhovad2021}, the multistable snap-through dynamics of mechanical metamaterials~\cite{Liu2023}, and the rich modal interactions of nonlinear mechanical oscillator networks~\cite{Coulombe2017} all provide the high-dimensional, nonlinear, and history-dependent responses that RC theory requires~\cite{Dambre2012}.

Although PRC theory and software have both advanced in recent years, the practical workflow for developing and evaluating mechanical physical reservoirs remains fragmented. Existing tools typically address only one part of the pipeline: mechanics packages such as MERLIN and PyElastica provide simulation capabilities for particular classes of structures, while PRC-oriented packages such as PRCpy focus primarily on preprocessing, training, and evaluation of reservoir data. What is still missing is a unified framework that treats both simulated trajectories and real experimental measurements through the same data interface, so that reservoir benchmarking, task-independent characterization, and design-space exploration can be carried out reproducibly across substrates and data sources. This gap makes it difficult to compare results fairly, reuse workflows between simulation and experiment, and systematically relate physical design choices to computational performance.

Domain-specific simulation tools have served individual subfields well. \textsc{MERLIN} and its successor \textsc{MERLIN2}~\cite{Liu2016,Liu2017,Liu2018} implement nonlinear bar-and-hinge models for quasi-static analysis of non-rigid origami, enabling efficient prediction of large-displacement and large-rotation responses critical to multistable origami mechanics~\cite{Filipov2017,Zhu2022}. PyElastica~\cite{Naughton2021,Gazzola2018} provides a modular Python simulator for assemblies of slender Cosserat rods, supporting biological and soft-robotic studies with full bend, twist, shear, and stretch degrees of freedom. ReservoirPy~\cite{Trouvain2020} offers a flexible interface for designing and benchmarking digital Echo State Networks. A closely related concurrent effort, OpenReservoirComputing (ORC)~\cite{Williams2026}, provides a GPU-accelerated JAX-based framework for digital reservoir computing with modular support for forecasting, classification, and control. Emerging tools such as PRCpy~\cite{Youel2024} begin to bridge physical systems and RC evaluation. However, none of these tools provide a common framework in which high-fidelity simulated trajectories and experimentally acquired data can be represented through the same schema and passed through the same downstream benchmarking, analysis, and optimization pipeline. Researchers therefore still need to stitch stages together using custom scripts, manual file conversions, and duplicated analysis code, making the workflow error-prone, difficult to reproduce, and difficult to scale.

To address this need, we introduce \textbf{OpenPRC}, an open-source Python framework for schema-driven physics-to-task evaluation and optimization in physical reservoir computing, built around three primary design objectives:

\begin{enumerate}[leftmargin=*, label=\textbf{\arabic*.}]
  \item \textbf{Unified workflow.} OpenPRC connects physical trajectory generation or ingestion, reservoir evaluation, analysis, and optimization within a single modular framework, reducing the overhead of coupling multiple disconnected software packages.

  \item \textbf{Schema-driven interoperability.} A five-module architecture (\texttt{demlat}, \texttt{openprc.vision}, \texttt{reservoir}, \texttt{analysis}, \texttt{optimize}) together with a universal HDF5 data schema enables both simulated trajectories and experimentally acquired data to enter the same downstream workflow in a reproducible and extensible format.

  \item \textbf{Physics-aware optimization.} By embedding physical governing equations directly into the optimization loop, OpenPRC enables the automated discovery of material parameters and structural topologies that maximize computational performance, directly linking physical design to information processing capability.
\end{enumerate}

The remainder of this paper is organized as follows. Section~\ref{sec:background} reviews the theoretical foundations of PRC and related software tools. Section~\ref{sec:architecture} describes OpenPRC's modular architecture and HDF5 schema. Sections~\ref{sec:demlat} and~\ref{sec:vision} detail the physics simulation engine and experimental ingestion layer respectively. Sections~\ref{sec:reservoir} and~\ref{sec:analysis} cover the learning and analysis modules. Section~\ref{sec:discussion} discusses limitations and future directions.

\section{Background and Related Work}
\label{sec:background}

\subsection{Reservoir computing fundamentals}
\label{subsec:rc_fundamentals}

A reservoir computer comprises three components~\cite{Jaeger2001,Maass2002}: (i)~an input layer mapping the driving signal $u(t)$ into the reservoir, (ii)~a fixed dynamical system whose state evolves as $\mathbf{x}(t+1)=f(\mathbf{W}\mathbf{x}(t)+\mathbf{W}_{\mathrm{in}}u(t))$ with fixed weights $\mathbf{W}$ and $\mathbf{W}_{in}$, (iii)~a trained linear readout $\hat{y}(t)=\mathbf{w}_{\mathrm{out}}^\top \mathbf{x}(t)$. The computational power of the reservoir depends on two key properties~\cite{Lukovsevivcius2009, Dambre2012}: \emph{fading memory}, which allows the system to retain information about recent inputs, and \emph{nonlinearity}, which allows it to construct higher-order transformations of those inputs.

\paragraph{Memory capacity.} Jaeger~\cite{Jaeger2002} introduced the memory capacity (MC), which measures a reservoir's ability to reconstruct delayed copies of its input:
\begin{equation}
    \mathrm{MC}=\sum_{k=1}^{\infty} C[u(t-k), \hat{y}_k(t)],
\end{equation}

with the bound $\mathrm{MC}\leq N$ for a reservoir with $N$ effective states. The per-delay capacity $C[u(t-k), \hat{y}_k(t)]$ is the squared Pearson correlation coefficient between the true delayed input $u(t-k)$ and the best linear readout output $\hat{y}_k(t)$ trained to reconstruct it:

\begin{equation}
C[u(t-k), \hat{y}_k(t)]
= \frac{\mathrm{Cov}^2\bigl(u(t-k),\,\hat{y}_k(t)\bigr)}
       {\sigma^2\bigl(u(t-k)\bigr)\,\sigma^2\bigl(\hat{y}_k(t)\bigr)}.
\label{eq:mc_capacity}
\end{equation}

\paragraph{Information Processing Capacity.} Dambre et al.~\cite{Dambre2012} generalized this through the framework of information processing capacity (IPC), which measures how well a reservoir can reconstruct a family of target functions of its input history. For a complete orthogonal basis of target functions $z$, the normalized capacity of each target is

\begin{equation}
    C[z]=1-\frac{\sigma^2(z-\hat{z})}{\sigma^2(z)},
    \qquad
    \sum_{z} C[z]\leq N,
\label{eq:dambre}
\end{equation}

where $\hat{z}$ is the best linear readout approximation of $z$. This provides a task-independent upper bound on computational capability: a reservoir with $N$ linearly independent observable states can realize at most $N$ independent functions of its input history. Note that when $z = u(t-k)$ and the basis is the set of all delays, the degree-one IPC sum reduces exactly to the classical memory capacity of Equation~\eqref{eq:mc_capacity}.

\subsection{Related software tools}
\label{subsec:software_gap}

Existing software tools address important parts of the PRC workflow but typically in isolation. Table~\ref{tab:software_comparison} summarizes the landscape. OpenPRC differs from all existing tools in providing a unified schema under which high-fidelity simulated trajectories, video-based experimental measurements, and data from third-party simulators can be benchmarked, analyzed, and optimized through the same pipeline.

\begin{table}[H]
\centering
\caption{Representative software tools related to PRC and their scope.}
\label{tab:software_comparison}
\footnotesize
\setlength{\tabcolsep}{5pt}
\renewcommand{\arraystretch}{1.3}
\begin{tabularx}{\linewidth}{@{} l p{3.2cm} >{\centering\arraybackslash}X
    >{\centering\arraybackslash}X >{\centering\arraybackslash}X
    >{\centering\arraybackslash}X @{}}
\toprule
\textbf{Tool} & \textbf{Primary scope} & \textbf{Physics sim.} &
\textbf{Experimental data} & \textbf{RC analysis} & \textbf{Optimization} \\
\midrule
MERLIN/MERLIN2   & Non-rigid origami mechanics        & $\checkmark$ & $\times$     & $\times$     & limited \\
PyElastica       & Cosserat rod simulation             & $\checkmark$ & $\times$     & $\times$     & limited \\
ReservoirPy      & Digital RC benchmarking             & $\times$     & $\times$     & $\checkmark$ & limited \\
ORC              & Digital RC in JAX                   & $\times$     & $\times$     & $\checkmark$ & limited \\
PRCpy            & PRC-oriented data workflow          & $\times$     & $\checkmark$ & $\checkmark$ & $\times$ \\
\textbf{OpenPRC} & \textbf{Unified physical PRC workflow} & $\checkmark$ & $\checkmark$ & $\checkmark$ & $\checkmark$ \\
\bottomrule
\end{tabularx}
\end{table}

\section{Framework Architecture}
\label{sec:architecture}

\subsection{Design philosophy}
\label{subsec:design}

OpenPRC is designed as a modular workflow for physical reservoir computing, guided by five practical principles:

\begin{enumerate}[leftmargin=*, label=\textbf{P\arabic*.}]
  \item \textbf{Separate physics from learning.} Simulation, ingestion, training, analysis, and optimization are handled by different modules so that each stage can be developed and used independently.

  \item \textbf{Use a shared data format.} Modules communicate through a common HDF5 schema, allowing simulated trajectories, video-derived measurements, and data from external simulators to enter the same downstream workflow.

  \item \textbf{Reproducible experiments.} OpenPRC stores simulation outputs, readout results, and evaluation metrics in a structured way so that the same experiment can be rerun, inspected, and compared consistently.

  \item \textbf{Stable interfaces across backends.} Physics models expose a common user-facing interface even when the underlying implementation differs, such as CPU or GPU execution.

  \item \textbf{File-based modular workflow.} Each module reads well-defined output files from the previous stage and writes its own results for the next, so users can reuse intermediate results without rerunning the entire pipeline.
\end{enumerate}

\vspace{3mm}
\subsection{Module overview}
\label{subsec:modules}

The framework comprises five modules arranged in a directed pipeline
(Figure~\ref{fig:pipeline}):

\begin{tcolorbox}[modulebox={demlat}{lightdemlat},
  title={\texttt{demlat} --- Discrete Element Modeling \& Lattice Simulation}]
High-fidelity simulation of nonlinear mechanical networks. Treats structures as graphs of discrete elements (nodes, bars, hinges, faces) and propagates their dynamics via a hybrid RK4--PBD solver. Produces trajectory tensors in \texttt{simulation.h5}, deliberately agnostic to RC concepts.
\end{tcolorbox}

\begin{tcolorbox}[modulebox={vision}{lightvision},
  title={\texttt{openprc.vision} --- Experimental Trajectory Ingestion}]
Converts raw video recordings into calibrated node-trajectory data in the same \texttt{simulation.h5} schema as \texttt{demlat}, via SIFT/ORB/AKAZE detection, pyramidal KLT tracking with forward-backward consistency, spatial calibration, and HDF5 serialization.
\end{tcolorbox}

\begin{tcolorbox}[modulebox={reservoir}{lightreservoir},
  title={\texttt{reservoir} --- Reservoir Computing Layer}]
Transforms physical trajectories into machine-learning-ready reservoir states. Applies user-defined feature extractors (node positions, displacements, bar-based features) and trains readout models through a common
\texttt{Trainer} interface.
\end{tcolorbox}

\begin{tcolorbox}[modulebox={analysis}{lightanalysis},
  title={\texttt{analysis} --- Characterization \& Benchmarking}]
Provides a statistical correlation toolkit (\texttt{openprc.analysis.correlation}) for inspecting reservoir structure, and benchmark wrappers for task-level and task-independent evaluation including NARMA, IPC decomposition, and user-defined benchmarks.
\end{tcolorbox}

\begin{tcolorbox}[modulebox={optimize}{lightoptimize},
  title={\texttt{optimize} --- Physics-Aware Optimization \textit{(in development)}}]
Closes the loop back to \texttt{demlat} for iterative structural and material optimization, enabling automated discovery of topologies and parameters that maximize information processing capacity.
\end{tcolorbox}

\subsection{Pipeline and data flow}
\label{subsec:pipeline}
\vspace{-4pt}
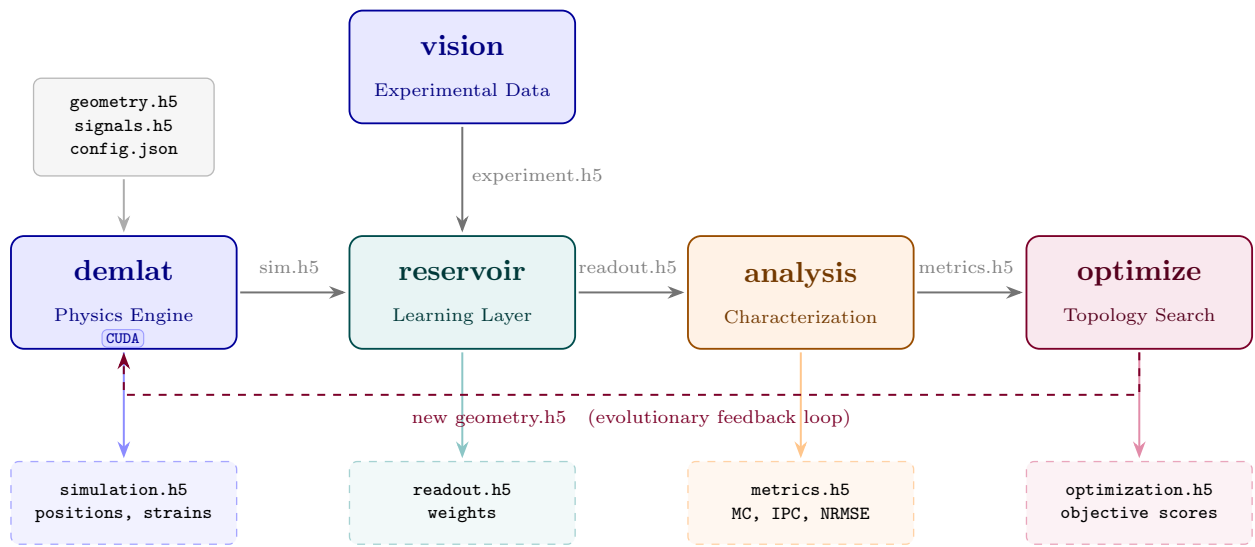
\begin{figure}[H]
\centering
\begin{center}
\begin{tikzpicture}[
  font=\small,
  >=Stealth,
  mod/.style={
    draw, rounded corners=5pt,
    minimum width=30mm, minimum height=15mm,
    align=center, line width=0.7pt,
    font=\small\bfseries
  },
  art/.style={
    draw, dashed, rounded corners=3pt,
    minimum width=30mm, minimum height=11mm,
    align=center, font=\ttfamily\fontsize{7.5}{9}\selectfont,
    line width=0.5pt
  },
  inp/.style={
    draw, rounded corners=3pt,
    minimum width=24mm, minimum height=13mm,
    align=center, font=\ttfamily\fontsize{7.5}{9}\selectfont,
    fill=gray!7, draw=gray!55, line width=0.5pt
  },
  lbl/.style={font=\fontsize{7.5}{9}\selectfont, align=center},
  arr/.style ={->, line width=0.75pt, shorten >=1pt, shorten <=1pt},
  arrd/.style={->, line width=0.75pt, dashed, shorten >=1pt, shorten <=1pt},
]

\def\dy{0}   
\node[mod, fill=blue!9,   draw=blue!60!black,  text=blue!50!black]
  (D) at (  0mm, \dy mm)
  {demlat\\[2pt]{\fontsize{7.5}{9}\selectfont\normalfont Physics Engine}};

\node[mod, fill=blue!9,   draw=blue!60!black,  text=blue!50!black]
  (V) at (  45mm, 3\dy mm)
  {vision\\[2pt]{\fontsize{7.5}{9}\selectfont\normalfont Experimental Data}};

\node[mod, fill=teal!9,   draw=teal!60!black,  text=teal!45!black]
  (R) at ( 45mm, \dy mm)
  {reservoir\\[2pt]{\fontsize{7.5}{9}\selectfont\normalfont Learning Layer}};

\node[mod, fill=orange!10, draw=orange!60!black, text=orange!45!black]
  (A) at ( 90mm, \dy mm)
  {analysis\\[2pt]{\fontsize{7.5}{9}\selectfont\normalfont Characterization}};

\node[mod, fill=purple!9,  draw=purple!65!black, text=purple!45!black]
  (O) at (135mm, \dy mm)
  {optimize\\[2pt]{\fontsize{7.5}{9}\selectfont\normalfont Topology Search}};

\node[inp] (INP) at (0mm, 22mm)
  {geometry.h5\\ signals.h5\\ config.json};

\draw[arr, gray!65] (INP.south) -- (D.north);

\node[draw=blue!40, fill=blue!13, rounded corners=2pt,
      font=\bfseries\fontsize{6}{7}\selectfont\ttfamily, text=blue!60!black,
      inner sep=1.5pt]
  at ($(D.south west)+(15.0mm, 1.5mm)$) {CUDA};

\draw[arr, black!55] (D) -- (R);
\node[lbl, text=black!50] at (22mm,  3.5mm) {sim.h5};

\draw[arr, black!55] (V) -- (R);
\node[lbl, text=black!50] at (55mm,  15.5mm) {experiment.h5};

\draw[arr, black!55] (R) -- (A);
\node[lbl, text=black!50] at (67mm,  3.5mm) {readout.h5};

\draw[arr, black!55] (A) -- (O);
\node[lbl, text=black!50] at (112mm,  3.5mm) {metrics.h5};

\node[art, fill=blue!5,    draw=blue!35]
  (SA) at (  0mm, -28mm)
  {simulation.h5\\ \upshape positions, strains};

\node[art, fill=teal!5,    draw=teal!35]
  (RA) at ( 45mm, -28mm)
  {readout.h5\\ \upshape weights};

\node[art, fill=orange!6,  draw=orange!35]
  (MA) at ( 90mm, -28mm)
  {metrics.h5\\ \upshape MC, IPC, NRMSE};

\node[art, fill=purple!5,  draw=purple!35]
  (OA) at (135mm, -28mm)
  {optimization.h5\\ \upshape objective scores};

\draw[arr, blue!45]   (D) -- (SA);
\draw[arr, teal!45]   (R) -- (RA);
\draw[arr, orange!45] (A) -- (MA);
\draw[arr, purple!45] (O) -- (OA);

\node[lbl, text=blue!65!black]   at (  0mm, -32mm) {} ;
\node[lbl, text=teal!60!black]   at ( 38mm, -32mm) {};
\node[lbl, text=orange!65!black] at ( 76mm, -32mm) {};
\node[lbl, text=purple!60!black] at (114mm, -32mm) {};

\draw[arrd, purple!65!black]
  (O.south) -- ++(0,-6mm)
             -| node[pos=0.25, below, lbl, yshift=-1pt]
                  {new geometry.h5\quad(evolutionary feedback loop)}
             (D.south);

\end{tikzpicture}
\end{center}
\caption{The OpenPRC pipeline. Both \texttt{demlat} and
\texttt{openprc.vision} produce \texttt{simulation.h5}, making simulated and experimental trajectories interchangeable to all downstream modules. The \texttt{optimize} module closes the loop back to \texttt{demlat} for iterative design optimization.}
\label{fig:pipeline}
\end{figure}
\vspace{-4pt}
Figure~\ref{fig:pipeline} illustrates the data flow. Each module reads the output file produced by the previous stage, and researchers can enter at any stage --- for example, by converting a third-party simulator output to the OpenPRC schema and passing it directly to \texttt{reservoir} and
\texttt{analysis}.

\subsection{Universal HDF5 schema}
\label{subsec:schema}

Interoperability is enforced through a strict schema hierarchy. The core input and output files in the workflow, together with their producers and consumers, are summarized in Table~\ref{tab:schema}.

\begin{table}[H]
\centering
\caption{Universal HDF5 file schema in OpenPRC.}
\label{tab:schema}
\footnotesize
\renewcommand{\arraystretch}{1.2}
\begin{tabularx}{\linewidth}{lllX}
\toprule
\textbf{File} & \textbf{Producer} & \textbf{Consumers} & \textbf{Key content} \\
\midrule
\texttt{geometry.h5} & User / \texttt{optimize} & \texttt{demlat}
  & Node coordinates $[N\times 3]$, masses $[N]$, bar connectivity $[M\times 2]$,
    hinge connectivity $[K\times 4]$, stiffness, rest lengths, damping \\
\texttt{signals.h5} & User / \texttt{optimize} & \texttt{demlat}
  & Named time series $[T_u]$ or $[T_u\times d]$; signal timestep \\
\texttt{config.json} & User / \texttt{optimize} & \texttt{demlat}
  & Simulation duration, integrator settings, actuator wiring, export options \\
\texttt{simulation.h5} & \texttt{demlat} / \texttt{openprc.vision} &
  \texttt{reservoir}, \texttt{analysis}
  & Node positions $[T_s\times N\times 3]$, velocities, bar strains
    $[T_s\times M]$, hinge angles $[T_s\times K]$, energies \\
\texttt{readout.h5} & \texttt{reservoir} & \texttt{analysis}, \texttt{optimize}
  & Readout weights $[D\times O]$, predictions $[T_s\times O]$, NMSE, NRMSE \\
\texttt{metrics.h5} & \texttt{analysis} & \texttt{optimize}, User
  & Memory-capacity profiles, IPC matrix, PCA variance, kernel rank \\
\texttt{optimization.h5} & \texttt{optimize} & User & Fitness history \\
\bottomrule
\end{tabularx}
\vspace{0.4em}
\begin{minipage}{\linewidth}
\scriptsize \textit{Notation:} $N$ nodes, $M$ bars, $K$ hinges, $T_u$
input-signal samples, $T_s$ saved simulation samples, $d$ signal dimension,
$D$ feature dimension, $O$ output dimension. Both \texttt{demlat} and
\texttt{openprc.vision} write to \texttt{simulation.h5}, making simulated and
experimental trajectories interchangeable to all downstream consumers.
\end{minipage}
\end{table}

\section{Physics Simulation Engine: \texttt{openprc.demlat}}
\label{sec:demlat}

\texttt{demlat} is the physics engine of OpenPRC. Given a mechanical structure, a set of input signals, and simulation settings, it produces time-domain trajectories that feed the downstream learning and analysis pipeline. The structure's geometry, actuation signals, and solver configuration are defined through structured input files; \texttt{demlat} reads these files, integrates the equations of motion, and writes the resulting trajectories to \texttt{simulation.h5}. Because the simulation layer writes a fixed-schema output file and makes no assumptions about what comes next, it is fully decoupled from the reservoir, analysis, and optimization stages.

\subsection{Physical representation: the bar-hinge model}
\label{subsec:barhinge_model}

The default model path is \texttt{demlat.models.barhinge}, which discretizes a mechanical structure as a graph of nodes, bars, and hinges --- the same representation used in nonlinear origami mechanics \cite{Liu2016,Liu2017,Liu2018,Zhu2022}. Nodes carry mass and position degrees of freedom. The acceleration of each node $\mathbf{P}_i$ with equivalent mass $m_i$ is governed by the sum of all acting forces:

\begin{equation}
\frac{d^2\mathbf{P}_i}{dt^2} = \frac{1}{m_i}
\Bigl(
  \mathbf{F}_i^{\mathrm{ext}}
  + \mathbf{F}_i^{\mathrm{axial}}
  + \mathbf{F}_i^{\mathrm{hinge}}
  + \mathbf{F}_i^{\mathrm{damping}}
\Bigr).
\label{eq:nodal_eom}
\end{equation}

Bars are two-node elements that resist axial deformation through a linear spring-damper law. Given axial stiffness $k_{\mathrm{axial}}$ and rest length $l_{ij}^0$, the axial force on node $i$ from its neighbours is

\begin{equation}
\mathbf{F}_i^{\mathrm{axial}}
= -k_{\mathrm{axial}}
  \sum_{j \in \mathcal{N}(i)}
  \bigl(l_{ij} - l_{ij}^0\bigr)\frac{\mathbf{r}_{ij}}{l_{ij}},
\label{eq:axial_force}
\end{equation}

and the corresponding viscous damping force is

\begin{equation}
\mathbf{F}_i^{\mathrm{damping}}
= -2\zeta\sqrt{m_i k_{\mathrm{axial}}}
  \sum_{j \in \mathcal{N}(i)}
  \Bigl(\dot{\mathbf{P}}_i - \dot{\mathbf{P}}_j\Bigr).
\label{eq:damping_force}
\end{equation}

Hinges are four-node elements defined by two triangular faces sharing a common edge; they resist changes in the dihedral angle $\theta$ relative to a prescribed rest angle $\theta_0$ through a restoring torque. For a hinge with stiffness $k_{\mathrm{hinge}}$ (set independently for fold lines and facet diagonals), the force on node $i$ is

\begin{equation}
\mathbf{F}_i^{\mathrm{hinge}}
= -k_{\mathrm{hinge}}\,(\theta - \theta_0)
  \frac{\partial\theta}{\partial\mathbf{P}_i},
\label{eq:hinge_force}
\end{equation}

where the angle gradients $\partial\theta/\partial\mathbf{P}_i$ are computed from the cross-product normals of the two adjacent faces \cite{Liu2017}. A key property of this formulation is that the acceleration of each node depends only on the positions and velocities of its immediate neighbours, which makes the force evaluation embarrassingly parallel and well-suited to GPU execution.

\begin{figure}[H]
    \centering
    \includegraphics[width=0.5\linewidth]{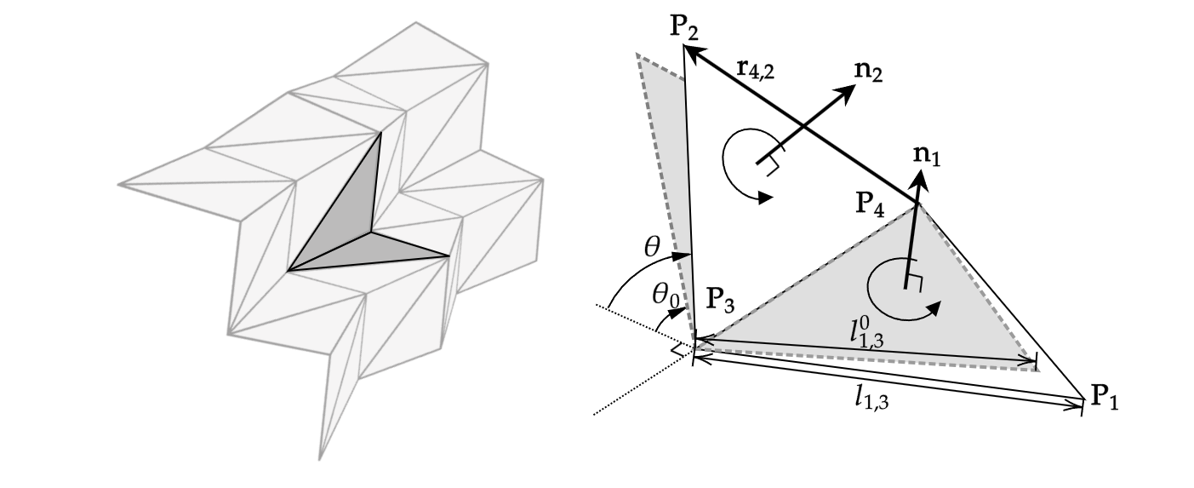}
    \caption{Bar-hinge element showing the four-node hinge configuration. Node $\mathbf{P}_i$ interacts with its neighbours through axial bar forces and dihedral hinge torques, with angle gradients computed from the normals of the two adjacent faces.} 
    \label{fig:barhinge_element}
\end{figure}

\begin{figure}[H]
    \centering
    \includegraphics[width=0.6\linewidth]{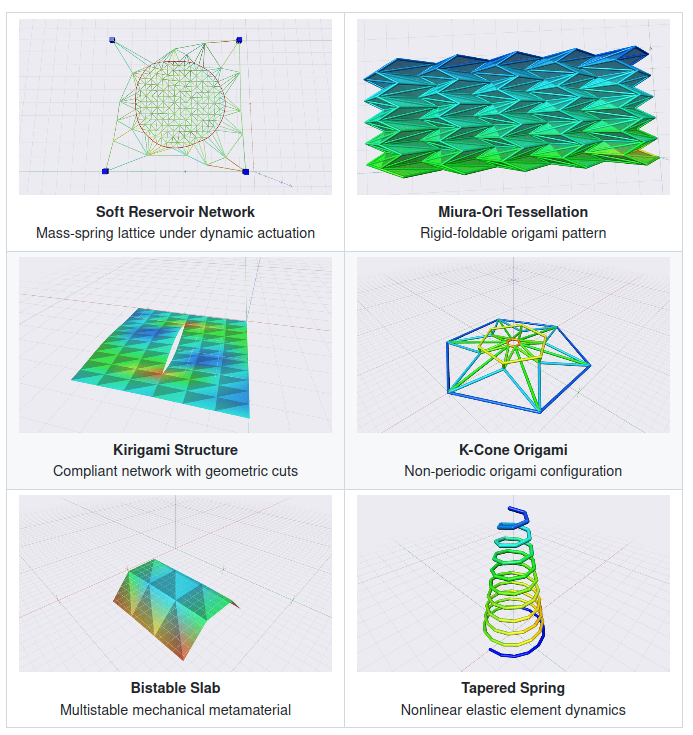}
    \caption{Versatile applications of the bar-hinge model in \texttt{demlat}: soft reservoir network (mass-spring lattice under dynamic actuation), Miura-ori tessellation, kirigami structure with geometric cuts, k-cone origami, bistable slab, and tapered spring. Strain is visualized using a cool--warm colormap.} \label{fig:barhinge_applications}
\end{figure}

Before integration begins, all geometric and material quantities are passed through a \texttt{SimulationScaler} that normalizes them to a consistent numerical scale, improving conditioning across the range of stiffness and mass values encountered in practice. Bars and hinges can be independently designated as soft (force-based) or rigid (constraint-based), making it straightforward to mix compliant fold lines with inextensible facet edges in the same model. By adjusting bar and hinge connectivity and stiffness assignments, \texttt{demlat.models.barhinge} extends naturally to a broad range of mechanical substrates beyond origami, as illustrated in Figure~\ref{fig:barhinge_applications}.

\subsection{Hybrid RK4--PBD solver}
\label{subsec:solver}

The numerical core of \texttt{demlat} is a hybrid integrator that couples explicit Runge--Kutta force advancement with a Position-Based Dynamics (PBD) projection stage \cite{Muller2007}. One simulation step can be written as the composed map

\begin{equation}
(\mathbf{x}^{n+1},\,\mathbf{v}^{n+1})
=
\mathcal{P}_{\mathrm{barhinge}}
\!\Bigl(
  \mathcal{R}_{\mathrm{RK4}}
  \!\bigl(
    \mathbf{x}^{n},\,\mathbf{v}^{n},\;
    \mathbf{f}^{\mathrm{bar}}
    +\mathbf{f}^{\mathrm{hinge}}
    +\mathbf{f}^{\mathrm{act}}
    +\mathbf{f}^{\mathrm{global}},\;
    \Delta t
  \bigr)
\Bigr),
\label{eq:demlat_split}
\end{equation}

where $\mathcal{R}_{\mathrm{RK4}}$ is the fourth-order Runge--Kutta update and $\mathcal{P}_{\mathrm{barhinge}}$ is the subsequent constraint projection.

During the RK4 stage, bar spring-damper forces, hinge restoring torques, actuator forces or prescribed displacements, and global forces such as gravity and viscous damping are evaluated at each sub-step and used to advance nodal positions and velocities. Soft bars and hinges are handled entirely at this stage through their constitutive force laws.

The projection stage then corrects any residual constraint violations introduced by the explicit integration. Rigid bars --- whose lengths must remain fixed --- are projected back onto the length constraint by iteratively displacing the two endpoint nodes along their connecting axis. Rigid or angle-restricted hinges are handled analogously by correcting the dihedral angle. This PBD loop iterates until all constraint errors fall below a specified tolerance, after which collision response is applied if enabled.

The hybrid design is motivated by practical considerations in mechanical reservoir computing. Soft elements such as fold-line hinges and axially compliant bars are most naturally described through continuous force laws, which the RK4 integrator handles efficiently at moderate stiffness values. Inextensible facet edges and rigid structural members, by contrast, would require prohibitively small timesteps if treated as very stiff springs; the PBD projection enforces these constraints exactly at each step regardless of their stiffness ratio. The
result is a solver that can represent origami, kirigami, tensegrity, and general compliant-mechanism geometries within a single unified time-stepping loop.

\subsection{Backend, execution, and user workflow}
\label{subsec:api}

\texttt{demlat} is backend-agnostic at the model level. The same \texttt{barhinge} model can be executed on CPU, via CUDA for GPU acceleration, or via JAX, with the backend selected at runtime. In automatic mode the implementation attempts CUDA first, then JAX, and falls back to CPU if neither is available. Simulation execution is managed by \texttt{Engine}, which validates the experiment directory, loads all simulation parameters and actuator wiring from the input files, runs the main physics loop, and writes the trajectory to \texttt{simulation.h5}. A post-processing pass appends derived quantities such as bar strains, hinge angles, and system energies after the main loop completes.

At the user level, the workflow is built around \texttt{SimulationSetup}, which provides a Python API for constructing the geometry, signals, and actuator wiring without directly writing HDF5 or JSON files. The following listing shows the key steps using a standing Miura-ori tessellation \cite{Bhovad2021,Filipov2017} as a representative substrate.

\begin{codeblock}[H]
\begin{lstlisting}
from openprc import demlat
from openprc.demlat import SimulationSetup, BarHingeModel

# --- simulation settings ---
setup = SimulationSetup("./experiments/miura_ori_4x8", overwrite=True)
setup.set_simulation_params(duration=10.0, dt=0.001, save_interval=0.01)
setup.set_physics(gravity=0, damping=0.8, enable_collision=True)

# --- geometry: nodes, bars, hinges ---
setup.add_node(position, mass=0.01)
setup.add_bar(node_i, node_j, stiffness=222.15, rest_length=L, damping=0.01)
setup.add_hinge([n0, n1, n2, n3], stiffness=0.01, rest_angle=theta)

# --- actuation: signal + actuator wiring ---
setup.add_signal("sig_base", signal_array, dt=0.0005)
setup.add_actuator(node_idx, "sig_base", type="position")

setup.save()

# --- run (CUDA backend) ---
engine = demlat.Engine(BarHingeModel, backend="cuda")
engine.run(demlat.Simulation("./experiments/miura_ori_4x8"))
\end{lstlisting}
\caption{\texttt{openprc.demlat} workflow showing geometry construction, actuation wiring, and simulation execution.}
\label{lst:demlat_miura}
\end{codeblock}

\begin{figure}[H]
    \centering
    \includegraphics[width=\linewidth]{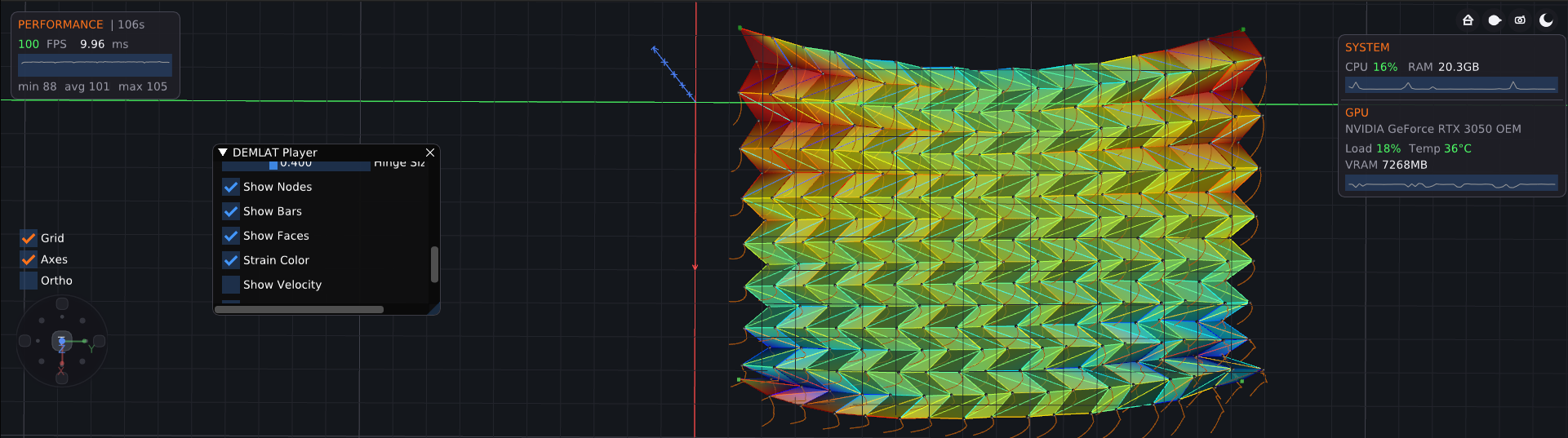}
    \caption{DEMLAT Player visualizing a Miura-ori simulation. Faces are colored by axial bar strain (cool--warm colormap), nodes and bars are overlaid, and node trajectory trails show the dynamic response under base excitation. The player is built on PiViz~\cite{phalak2026piviz}, a high-performance GPU-accelerated visualization library developed alongside OpenPRC.}
    \label{fig:miura_demlat_demo}
\end{figure}

Although the example uses simulated trajectories, the downstream modules impose no such restriction. Any experimental trajectory stored in the standardized \texttt{simulation.h5} schema --- for instance, node displacements extracted from high-speed video via \texttt{openprc.vision} --- can be passed directly to \texttt{reservoir} and \texttt{analysis} without modifying a single line of the downstream code.

\section{Experimental Trajectory Ingestion: \texttt{openprc.vision}}
\label{sec:vision}

\texttt{openprc.vision} is the experimental data ingestion module of OpenPRC. Its role is to convert raw video recordings of a physical reservoir into the same calibrated node-trajectory format that \texttt{demlat} produces from simulation, so that the downstream \texttt{reservoir} and \texttt{analysis} stages operate identically regardless of whether the data originated from a GPU simulation or a laboratory camera. The module implements a self-contained pipeline: video ingestion, keypoint detection with optional sidecar caching, KLT optical-flow tracking, spatial calibration, trajectory post-processing, and HDF5 serialization. All outputs are written to the same \texttt{simulation.h5} schema described in Section~\ref{sec:architecture}, making \texttt{openprc.vision} the primary bridge between physical experiment and the rest of the framework.

\subsection{Feature detection}
\label{subsec:vision_detection}

The first stage identifies a set of candidate tracking points on a reference frame. \texttt{openprc.vision} provides three built-in detectors --- SIFT, ORB, and AKAZE --- together with a registration interface for user-defined detectors.

SIFT \cite{Lowe2004} detects keypoints as local extrema of a Difference-of-Gaussians scale space and describes each point with a 128-dimensional histogram of oriented gradients, giving it strong invariance to scale, rotation, and moderate illumination change. It is the recommended default for laboratory recordings where lighting conditions are reasonably controlled and tracking density matters more than speed. ORB \cite{Rublee2011} combines the FAST corner detector with a binary BRIEF descriptor, reducing descriptor dimensionality to 32 bytes and making detection substantially faster at the cost of some robustness. AKAZE \cite{Alcantarilla2013} operates in a nonlinear scale space built from the Modified-Linear Diffusion equation, which preserves edges more faithfully than Gaussian blurring and often produces more stable keypoints on structured surfaces such as origami crease patterns.

All three detectors share the same callable interface and return a \texttt{FeatureMap} object storing keypoint coordinates, descriptors, response strengths, scales, and orientations. A \texttt{FeatureMapCache} companion class writes detected feature maps to a sidecar HDF5 file adjacent to the video and validates them against the video content hash on subsequent runs, so that detection is skipped automatically when the reference frame has not changed. The top-$N$ features by response strength can be selected via \texttt{FeatureMap.top\_n()} before tracking begins, which reduces downstream computation when the full detected set is larger than needed.

\subsection{KLT tracking and forward-backward consistency}
\label{subsec:vision_tracking}

Feature positions are tracked through the video using the Lucas--Kanade optical flow estimator \cite{LucasKanade1981} in its pyramidal formulation \cite{Bouguet2001}. At each pyramid level, the method solves for the displacement $\mathbf{d}$ that minimizes the sum-of-squared photometric residual in a local window around each keypoint,

\begin{equation}
\mathbf{d}^* = \arg\min_{\mathbf{d}}
\sum_{\mathbf{p} \in \mathcal{W}}
\bigl[I_1(\mathbf{p}) - I_2(\mathbf{p} + \mathbf{d})\bigr]^2,
\label{eq:klt}
\end{equation}

where $I_1$ and $I_2$ are consecutive frames, $\mathcal{W}$ is the search window, and the minimization is linearized via the image gradient (the standard brightness-constancy assumption). The pyramidal extension tracks large displacements by solving the flow problem coarse-to-fine across a Gaussian image pyramid \cite{Bouguet2001}, with the number of pyramid levels and the per-level window size exposed as \texttt{KLTConfig} parameters.

To suppress drift on features that have genuinely left the frame or become occluded, \texttt{openprc.vision} applies a forward-backward consistency check \cite{Kalal2010} after each frame pair. The feature is tracked forward from frame $t$ to $t+1$ and then backward from $t+1$ to $t$; if the Euclidean distance between the original position and the round-trip position exceeds a configurable threshold, the feature is marked as \texttt{DRIFT} and excluded from subsequent tracking. Features that move outside the frame boundary are marked \texttt{OOB}. The resulting \texttt{TrackingResult} stores a $(T \times N \times 2)$ position tensor and a corresponding status tensor using integer codes (\texttt{TRACKED}, \texttt{LOST}, \texttt{OOB}, \texttt{DRIFT}, \texttt{INTERPOLATED}, \texttt{REFERENCE}), making it straightforward to inspect tracking quality on a per-feature, per-frame basis.

\subsection{Calibration, trajectory extraction, and user workflow}
\label{subsec:vision_workflow}

Raw pixel coordinates are converted to physical units through a calibration object before the trajectory set is constructed. \texttt{NormalizedCalibration} maps pixel coordinates linearly to the unit square and requires no camera parameters, serving as a zero-configuration default. \texttt{CameraCalibration} implements the full pinhole model with optional radial and tangential lens distortion, accepting either a $3 \times 3$ intrinsic matrix and distortion coefficients from a standard OpenCV calibration procedure \cite{Zhang2000,Bradski2000}, or a physical scale factor that converts normalized camera coordinates directly to millimetres or another unit of length.

The calibrated positions are held in a \texttt{TrajectorySet} object with shape $(T \times N \times 2)$. Short tracking gaps can be filled by linear interpolation up to a configurable maximum gap length, and a NaN-aware moving average smoother suppresses sub-pixel jitter before export. From the trajectory set, one-dimensional, two-dimensional, or three-dimensional signal arrays are extracted via \texttt{to\_signals()}, selecting the $x$ component, $y$ component, displacement magnitude from the reference position, or cumulative path length as needed. Velocity and speed tensors are also available as derived quantities. The full pipeline state --- video metadata, feature map, tracking result, trajectory set, and signal array --- is serialized to a single HDF5 file through \texttt{openprc.vision.save()}, and any combination of components can be reloaded independently via \texttt{openprc.vision.load()}.

The following listing shows the key steps of the pipeline applied to an experimental video recording of a physical Miura-ori reservoir \cite{Bhovad2021}.

\begin{codeblock}[H]
\begin{lstlisting}
import openprc.vision as vision

# --- open video, detect and cache features ---
src   = vision.VideoSource("experiment.mp4")
fmap  = vision.FeatureMapCache(src).get_or_detect(
            src.read_frame(0), method="sift", contrast_threshold=0.04)
fmap  = fmap.top_n(200)

# --- track with forward-backward consistency check ---
src.reset()
result = vision.Tracker(
             src, fmap,
             config=vision.KLTConfig(win_size=(21,21),
                                     max_level=3,
                                     fb_threshold=1.0)).run()

# --- calibrate and build trajectory set ---
cal  = vision.CameraCalibration.from_matrix(
           K, frame_shape=(1080, 1920),
           scale_factor=25.0, physical_units="mm")
tset = vision.TrajectorySet.from_tracking_result(result, calibration=cal)

# --- filter, fill gaps, smooth ---
good = result.track_ratios() >= 0.6
tset = tset.select_features(good).fill_gaps("linear", max_gap=10).smooth(5)

# --- extract signals and save ---
vision.save("experiment_vision.h5", source=src, feature_map=fmap,
            tracking_result=result, trajectory_set=tset,
            signals=tset.to_signals(dim=2),
            signals_meta={"dim": 2, "units": tset.units})
src.release()
\end{lstlisting}
\caption{\texttt{openprc.vision} pipeline from video to calibrated signal array: feature detection, KLT tracking, calibration, and HDF5 serialization.}
\label{lst:vision_pipeline}
\end{codeblock}

\begin{figure}[H]
    \centering
    \includegraphics[width=0.6\linewidth]{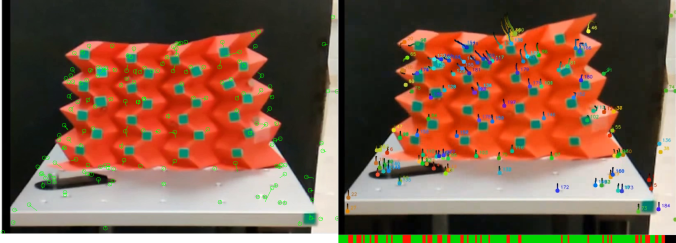}
    \caption{Output of the \texttt{openprc.vision} pipeline applied to a physical Miura-ori reservoir \cite{Bhovad2021}. Left: SIFT keypoints detected on the reference frame, with feature indices and response circles overlaid. Right: KLT tracking in progress, showing per-feature trajectory trails and forward-backward status indicators. The status bar at the bottom encodes per-feature tracking health across the frame window.}
    \label{fig:vision_miura}
\end{figure}

The output file \texttt{experiment\_vision.h5} follows the same universal HDF5 schema as \texttt{simulation.h5}, and the extracted signal array can be passed directly to \texttt{StateLoader} in the \texttt{reservoir} module without any further conversion. This makes \texttt{openprc.vision} the practical completion of the simulation-to-experiment interoperability claim: a researcher can switch between a CUDA simulation and a camera recording by changing a single file path.

\section{Learning Layer: \texttt{openprc.reservoir}}
\label{sec:reservoir}

The \texttt{reservoir} module is the learning layer of OpenPRC. Its role is to convert saved physical trajectories into reservoir-state features and fit a readout model for a downstream task. The layer is organized around four pieces: trajectory loading, feature extraction, readout model selection, and supervised training.

Trajectory access is provided by \texttt{StateLoader}, which reads \texttt{simulation.h5} and exposes helper methods for node positions, node displacements, bar lengths, bar extensions, and actuation signals. Feature construction is handled by dedicated feature objects such as \texttt{NodePositions}, \texttt{NodeDisplacements}, \texttt{BarLengths}, and \texttt{BarExtensions}, which transform the physical trajectory into a design matrix suitable for readout training. Readout fitting is handled by \texttt{Trainer}, which combines a trajectory loader, a feature extractor, and a readout model such as \texttt{Ridge}, applies a fixed washout period and train/test split, and fits the readout on a supplied task target.

The downstream modules operate directly on the saved \texttt{simulation.h5} produced by \texttt{demlat}, without rerunning the physics simulation. This interface is not limited to numerically generated data: if experimentally measured trajectories are stored in the same standardized \texttt{simulation.h5} format --- whether produced by \texttt{demlat} or ingested from video by \texttt{openprc.vision} --- the same feature extraction, readout training, benchmark evaluation, and visualization routines apply unchanged.

\begin{codeblock}[H]
\begin{lstlisting}
from openprc.reservoir.io.state_loader import StateLoader
from openprc.reservoir.features.node_features import NodeDisplacements
from openprc.reservoir.readout.ridge import Ridge
from openprc.reservoir.training.trainer import Trainer

loader  = StateLoader("./experiments/miura_ori_8x4/output/simulation.h5")
trainer = Trainer(
    loader=loader,
    features=NodeDisplacements(reference_node=0, dims=[2]),
    readout=Ridge(1e-5),
    experiment_dir="./experiments/miura_ori_8x4",
    washout=5.0, train_duration=10.0, test_duration=10.0,
)
\end{lstlisting}
\caption{Trajectory loading and trainer construction in
         \texttt{openprc.reservoir}.}
\label{lst:reservoir_basic}
\end{codeblock}

\section{Characterization and Benchmarking: \texttt{openprc.analysis}}
\label{sec:analysis}

The \texttt{analysis} module sits on top of the trained reservoir pipeline and provides two complementary diagnostic layers: a statistical correlation toolkit for inspecting the structure of reservoir dynamics before any task is defined, and a set of benchmark wrappers for evaluating task-level and task-independent computational performance. In practice, the correlation layer is the natural first stop after a simulation or experimental recording, while the benchmark layer quantifies what that reservoir can actually compute.

\subsection{Reservoir diagnostics: \texttt{openprc.analysis.correlation}}
\label{subsec:correlation}

Before committing to readout training, it is useful to ask three structural questions about the reservoir state matrix $X \in \mathbb{R}^{T \times N}$: how well each channel encodes the input signal, what nonlinear structure the reservoir is exploiting, and how much of the available dimensionality is genuinely independent. \texttt{openprc.analysis.correlation} answers these questions through three classes --- \texttt{Linear}, \texttt{Nonparametric}, and \texttt{Redundancy} --- all of which operate on the raw state matrix and return \texttt{Result} containers that support statistical testing, LaTeX export, and matplotlib plotting.

\paragraph{Linear input--reservoir relationships.}

\texttt{Linear(u, X)} computes the family of linear dependence metrics between a scalar input $u \in \mathbb{R}^T$ and the reservoir state matrix $X \in \mathbb{R}^{T \times N}$. The zero-lag Pearson correlation

\begin{equation}
r_i = \frac{\sum_{t=1}^{T}(u_t - \bar{u})(x_{it} - \bar{x}_i)}
           {\sqrt{\sum_{t}(u_t-\bar{u})^2 \sum_{t}(x_{it}-\bar{x}_i)^2}}
\label{eq:pearson}
\end{equation}

measures the instantaneous linear coupling between the input and channel $i$. Because physical reservoirs introduce latency through wave propagation and modal coupling, the cross-correlation function

\begin{equation}
\mathrm{CCF}_i(\tau)
= \frac{\sum_{t}(u_t - \bar{u})(x_{i,t+\tau} - \bar{x}_i)}
       {\sqrt{\sum_t(u_t-\bar{u})^2 \sum_t(x_{it}-\bar{x}_i)^2}}
\label{eq:ccf}
\end{equation}

is more informative in practice: its peak identifies the lag at which each channel is most responsive, and a spread of optimal lags across channels indicates that the reservoir is constructing a temporal basis of the input history. \texttt{Linear} also exposes the partial correlation matrix via the precision-matrix identity $\rho_{ij} = -P_{ij}/\sqrt{P_{ii}P_{jj}}$ where $P = \Sigma^{-1}$ and $\Sigma$ is the covariance matrix, which reveals direct channel-to-channel dependencies after removing the influence of all other channels. Canonical Correlation Analysis is available through \texttt{Linear.cca()}, which finds the optimal linear combination of reservoir channels that best reconstructs the input and provides an upper bound on what a linear readout can recover.

\paragraph{Nonlinear dependencies.}

Pearson and CCF are blind to nonlinear structure. A channel that encodes $x^2$, $|x|$, or $\sin(x)$ will appear uncorrelated under $r_i \approx 0$ even though it carries substantial information. \texttt{Nonparametric(x, y)} provides three complementary detectors for such structure.

Spearman's rank correlation $r_s$ and Kendall's $\tau_b$ capture monotonic but nonlinear relationships by operating on ranks rather than raw values. Distance correlation \cite{Szekely2007} is strictly stronger: given doubly centered pairwise distance matrices $A$ (from $x$) and $B$ (from channel $y_i$),

\begin{equation}
\mathrm{dCor}(x, y_i)
= \frac{\mathrm{dCov}(x,y_i)}
       {\sqrt{\mathrm{dCov}(x,x)\cdot\mathrm{dCov}(y_i,y_i)}},
\qquad
\mathrm{dCov}^2(x,y_i) = \frac{1}{T^2}\sum_{k,l}A_{kl}B_{kl},
\label{eq:dcor}
\end{equation}

and satisfies $\mathrm{dCor}(x,y_i) = 0$ if and only if $x$ and $y_i$ are statistically independent --- a property Pearson does not share. Channels with $|r_i| \approx 0$ but $\mathrm{dCor} \gg 0$ are nonlinearly encoding the input and will be invisible to a linear readout unless a feature expansion is applied. The Hilbert--Schmidt Independence Criterion (HSIC) \cite{Gretton2005} provides a kernel-based alternative,

\begin{equation}
\mathrm{HSIC}(x, y_i) = \frac{1}{T^2}\,\mathrm{tr}(KHLH),
\label{eq:hsic}
\end{equation}

where $K$ and $L$ are RBF kernel matrices with median-heuristic bandwidth and $H = I - T^{-1}\mathbf{1}\mathbf{1}^\top$ is the centering matrix. HSIC is sensitive to specific frequency structures through the kernel choice and serves as a useful cross-check against dCor on channels with suspected periodic or localized nonlinearity.

\paragraph{Internal redundancy.}

\texttt{Redundancy(y)} characterizes the internal correlation structure of the reservoir without reference to any input signal, answering whether all $N$ channels are genuinely contributing independent directions. The effective rank

\begin{equation}
r_{\mathrm{eff}}
= \exp\!\Bigl(-\sum_{i=1}^{N}\hat\lambda_i\ln\hat\lambda_i\Bigr),
\qquad
\hat\lambda_i = \frac{\lambda_i}{\sum_j\lambda_j},
\label{eq:effective_rank}
\end{equation}

is the exponential of the Shannon entropy of the normalized eigenvalue spectrum of the correlation matrix \cite{Roy2007}. When $r_{\mathrm{eff}} \approx N$ all eigenvalues are equal and every channel contributes an independent direction; when $r_{\mathrm{eff}} \approx 1$ a single mode dominates and the reservoir is effectively one-dimensional regardless of how many nodes are observed. The condition number $\kappa = \lambda_{\max}/\lambda_{\min}$ directly governs readout stability: when $\kappa$ is large, small noise in $Y$ causes large swings in the least-squares readout weights, and ridge regularization becomes essential. Single-linkage clustering on the absolute pairwise correlation matrix at a user-supplied threshold identifies groups of near-collinear channels; keeping one representative per group reduces $\kappa$ without sacrificing predictive power.

\paragraph{Usage.}

All three classes share a lazy-evaluation interface: metrics are computed on first access and cached, so calling \texttt{lin.pearson} twice incurs only one computation. Every metric returns a \texttt{Result} object that supports FDR-corrected and Bonferroni-corrected $p$-values via \texttt{Result.significant()}, DataFrame and \LaTeX\ export, and automatic plot dispatch (\texttt{"bar"} for per-channel scalars, \texttt{"heatmap"} for matrices, \texttt{"lag\_profile"} for CCF profiles).

\begin{codeblock}[H]
\begin{lstlisting}
from openprc.analysis import correlation as corr
from openprc.reservoir.io.state_loader import StateLoader
from openprc.reservoir.features.node_features import NodePositions

loader = StateLoader("./experiments/miura_ori_8x4/output/simulation.h5")
Y = NodePositions(dims=[2]).transform(loader)        # (T, N)
x = loader.get_actuation_signal(actuator_idx=0, dof=2)

# --- linear structure ---
lin = corr.Linear(x, Y, lag_sweep=True)
lin.pearson.plot()                   # per-channel Pearson r
lin.ccf.plot(kind="lag_profile")     # CCF with peak markers
lin.partial.plot()                   # N x N partial correlation heatmap
lin.cca()                            # best linear reconstruction of x

# --- nonlinear dependencies ---
corr.Nonparametric(x, Y).dcor.plot() # flag Pearson-blind channels

# --- redundancy ---
red = corr.Redundancy(Y)
print(f"Effective rank:  {red.rank:.1f} / {Y.shape[1]}")
print(f"Condition number: {red.condition:.2e}")
for g in red.groups_named(threshold=0.85):
    if len(g) > 1: print(f"  Redundant cluster: {g}")
\end{lstlisting}
\caption{Reservoir diagnostic workflow using
         \texttt{openprc.analysis.correlation}.}
\label{lst:correlation}
\end{codeblock}

\subsection{Task-level benchmark: NARMA}
\label{subsec:narma}

A first use of the \texttt{analysis} layer is to evaluate the reservoir on a standard task benchmark. Built-in tasks such as \texttt{NARMA\_task} are imported from \texttt{analysis.tasks.imitation}, benchmark wrappers such as \texttt{NARMABenchmark}, \texttt{MemoryBenchmark}, and \texttt{CustomBenchmark} from \texttt{analysis.benchmarks}, and plotting through \texttt{TimeSeriesComparison}. In the following example, the Miura-ori trajectories are evaluated through \texttt{NARMABenchmark}, which internally constructs the NARMA2 target from the stored base-excitation signal and reports the resulting prediction error.

\begin{codeblock}[H]
\begin{lstlisting}
from openprc.reservoir.io.state_loader import StateLoader
from openprc.reservoir.features.node_features import NodePositions
from openprc.reservoir.readout.ridge import Ridge
from openprc.reservoir.training.trainer import Trainer
from openprc.analysis.benchmarks.narma_benchmark import NARMABenchmark
from openprc.analysis.visualization.time_series import TimeSeriesComparison

loader  = StateLoader("./experiments/miura_ori_8x4/output/simulation.h5")
u_raw   = loader.get_actuation_signal(actuator_idx=0, dof=2)
u_input = 0.5 * (u_raw - u_raw.min()) / (u_raw.max() - u_raw.min())

trainer = Trainer(loader=loader, features=NodePositions(dims=[2]),
                  readout=Ridge(1e-5), experiment_dir="./experiments/miura_ori_8x4",
                  washout=5.0, train_duration=10.0, test_duration=10.0)

score = NARMABenchmark(group_name="narma_benchmark").run(
            trainer, u_input, order=2)
score.save()
TimeSeriesComparison().plot(score.readout_path, start_frame=0, end_frame=500).save()
\end{lstlisting}
\caption{NARMA2 benchmark using \texttt{NARMABenchmark}.}
\label{lst:narma}
\end{codeblock}

\begin{figure}[H]
    \centering
    \includegraphics[width=0.78\linewidth]{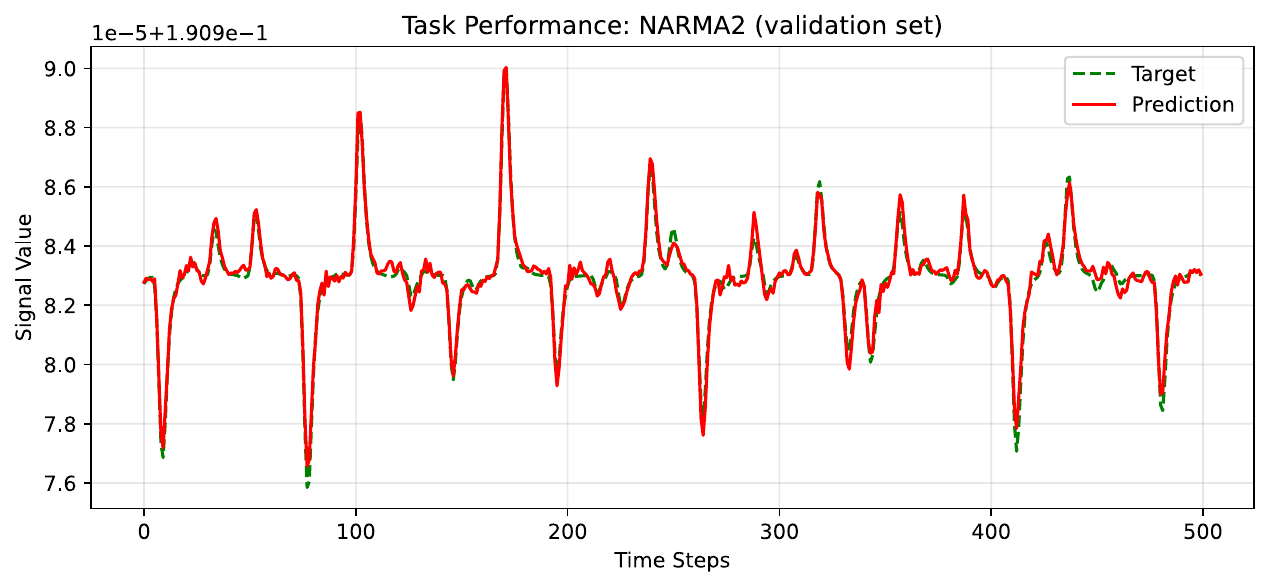}
    \caption{Representative NARMA2 benchmark result for the Miura-ori reservoir. The target is generated from the normalized base-excitation signal and the prediction is obtained from the trained readout over the evaluation window.}
    \label{fig:miura_narma2_timeseries}
\end{figure}

Figure~\ref{fig:miura_narma2_timeseries} provides a direct task-level view of performance by comparing the target and readout prediction over time. This complements the scalar benchmark metrics stored in \texttt{metrics.h5} and shows how accurately the physical reservoir reproduces the nonlinear target dynamics.

\paragraph{Task-independent diagnostic: memory and information-processing capacity.}

While NARMA2 summarizes performance on a particular nonlinear benchmark, it does not reveal \emph{why} a given physical reservoir performs well or poorly. OpenPRC therefore also supports a memory benchmark based on a truncated Dambre-style information-processing-capacity (IPC) decomposition \cite{Dambre2012}. Rather than evaluating only one task target, this benchmark constructs a delayed basis of target functions from the applied input sequence and fits a readout for each basis element, providing a structural picture of what information about the past input is retained by the reservoir.

For an applied input sequence $u(t)$, the benchmark first rescales to the Legendre domain,

\[
u_{\mathrm{leg}}(t)=2\frac{u(t)-u_{\min}}{u_{\max}-u_{\min}}-1,
\]

and constructs delayed basis functions over the lag set $\mathcal{L}=\{0,k_{\mathrm{delay}},2k_{\mathrm{delay}},\dots,\tau_s k_{\mathrm{delay}}\}$. For each exponent vector $\alpha=(\alpha_0,\dots,\alpha_{\tau_s})$ with total degree $1\le \sum_j \alpha_j \le n_s$, the target basis function is

\[
y_\alpha(t)=\prod_{j=0}^{\tau_s}\widetilde{P}_{\alpha_j}\!
\bigl(u_{\mathrm{leg}}(t-j\,k_{\mathrm{delay}})\bigr),
\]

where $\widetilde{P}_n(x)=\sqrt{2n+1}\,P_n(x)$ is the normalized Legendre polynomial of degree $n$. A ridge readout is fitted for each target and the raw capacity is

\[
C_{\mathrm{raw}}[\alpha]
=1-\frac{\sum_t (y_\alpha(t)-\hat y_\alpha(t))^2}
        {\sum_t (y_\alpha(t)-\bar y_\alpha)^2}.
\]

Capacities below a threshold $\epsilon$ are zeroed to suppress finite-data overestimation:

\[
C[\alpha]=
\begin{cases}
C_{\mathrm{raw}}[\alpha], & C_{\mathrm{raw}}[\alpha]>\epsilon,\\[4pt]
0, & C_{\mathrm{raw}}[\alpha]\le\epsilon.
\end{cases}
\]

The threshold is chosen from an effective-rank-aware chi-squared rule: for effective rank $N$ and test window length $T$,

\[
P\!\left(\chi^2(N)\ge t\right)=p, \qquad \epsilon=\frac{2t}{T},
\]

where the factor of $2$ is a conservative correction for the non-independence of physical reservoir states. The total IPC decomposes as

\[
\mathrm{MC}_{\mathrm{lin}}=\sum_{\deg(\alpha)=1} C[\alpha], \qquad
\mathrm{MC}_{\mathrm{nonlin}}=\sum_{\deg(\alpha)>1} C[\alpha], \qquad
\mathrm{IPC}_{\mathrm{tot}}=\sum_{\deg(\alpha)\ge 1} C[\alpha].
\]

When $k_{\mathrm{delay}}=1$, the degree-one component coincides with the standard Jaeger-style memory-capacity benchmark \cite{Jaeger2002}.

\begin{codeblock}[H]
\begin{lstlisting}
from scipy.stats import chi2
from sklearn.preprocessing import StandardScaler
from openprc.analysis.benchmarks.memory_benchmark import MemoryBenchmark

# --- effective rank and epsilon (reuse loader, features, trainer from above) ---
X   = StandardScaler().fit_transform(features.transform(loader))
s   = np.linalg.svd(X, compute_uv=False)
N   = float(np.exp(-np.sum((s/s.sum()) * np.log(s/s.sum() + 1e-12))))
eps = float(2 * chi2.isf(1e-4, df=N) / int(10.0 / loader.dt))

score = MemoryBenchmark(group_name="memory_benchmark").run(
            trainer, u_input,
            tau_s=30, n_s=2, k_delay=1, eps=eps, ridge=1e-6)
score.save()
\end{lstlisting}
\caption{Memory benchmark with effective-rank-based $\epsilon$ selection.}
\label{lst:memory}
\end{codeblock}

\begin{figure}[H]
    \centering
    \includegraphics[width=\linewidth]{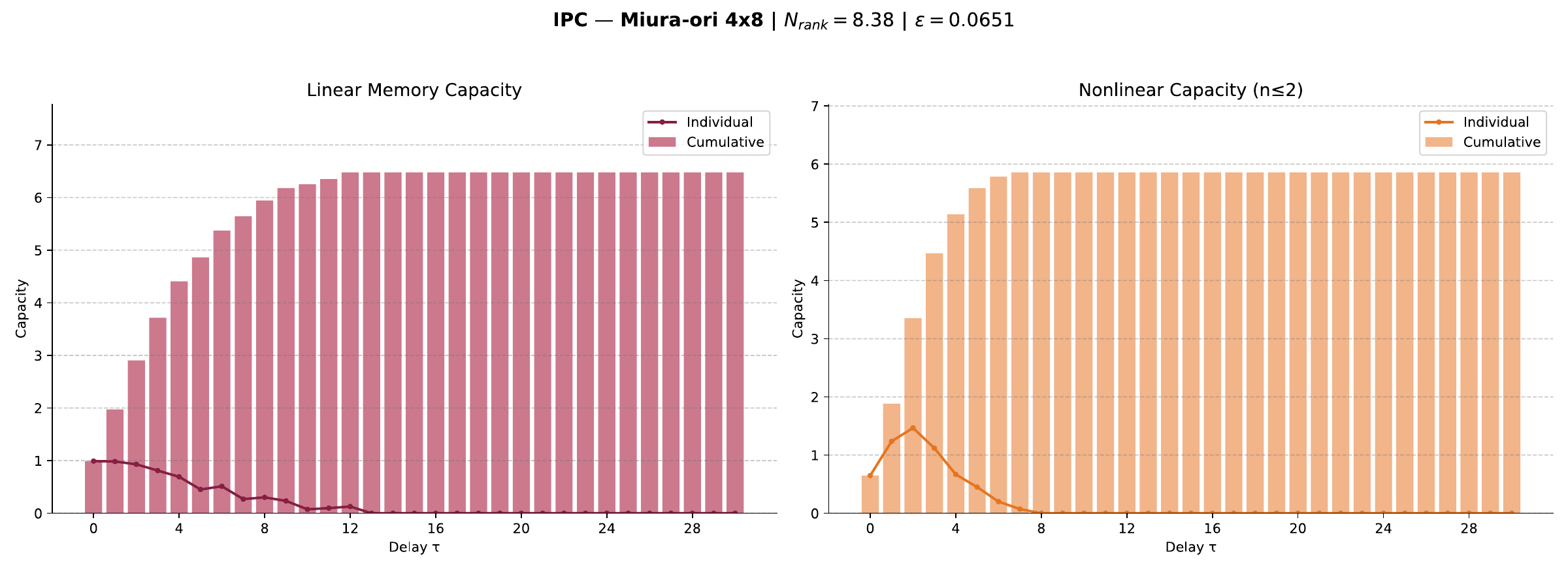}
    \caption{Memory-benchmark result for the Miura-ori reservoir under the truncated Dambre-style IPC analysis. Linear and nonlinear capacity contributions are resolved across delayed input components.}
    \label{fig:ipc_analysis}
\end{figure}

Figure~\ref{fig:ipc_analysis} complements the NARMA2 result by showing how the reservoir distributes retained capacity across delayed input components. Whereas the NARMA benchmark provides a task-level error summary, the IPC-style benchmark resolves which parts of the past input remain available to the readout and therefore provides a more interpretable picture of the reservoir's underlying computational structure.

This benchmark should be interpreted as a finite-data empirical approximation to Dambre-style IPC rather than an exact theoretical capacity measurement. The main practical limitations are that physical experiments rarely admit perfectly iid excitation, the basis search is necessarily truncated by $(\tau_{\max},n_{\max},k_{\mathrm{delay}})$, the $\epsilon$-threshold introduces a conservative downward bias, and the result depends on the chosen feature representation rather than the full hidden physical state.

\paragraph{Extensible benchmark logic.}

Beyond the built-in families, user-defined benchmark logic is supported through \texttt{CustomBenchmark}. The benchmark layer is extensible by design: any callable that accepts a trainer and input signal and returns a metrics dictionary can be registered as a benchmark.

\begin{codeblock}[H]
\begin{lstlisting}
from openprc.analysis.benchmarks.custom_benchmark import CustomBenchmark
from openprc.analysis.tasks.imitation import NARMA_task

def custom_narma_logic(benchmark, trainer, u_input, **kwargs):
    order  = kwargs.get("order", 2)
    result = trainer.train(NARMA_task(u_input, order=order),
                           task_name=f"CustomNARMA{order}")
    result.save()
    _, target, pred = result.cache["test"]
    nrmse = np.sqrt(np.mean((target - pred)**2)) / (np.std(target) or 1.0)
    return {f"custom_narma{order}_nrmse": nrmse}, {"narma_order": order}

score = CustomBenchmark(
            group_name="custom_narma_benchmark",
            benchmark_logic=custom_narma_logic,
        ).run(trainer, u_input, order=2)
score.save()
\end{lstlisting}
\caption{User-defined benchmark logic via \texttt{CustomBenchmark}.}
\label{lst:custom}
\end{codeblock}

\section{Implementation Status, Limitations, and Near-Term Roadmap}
\label{sec:discussion}

\subsection{Current implementation status}
\label{subsec:status}

OpenPRC is presented in this paper as an active software effort rather than a completed end-state platform. Among the five modules, \texttt{demlat} is the most mature and serves as the primary physics front end. \texttt{openprc.vision} provides a stable experimental ingestion path for video-based trajectory extraction. The \texttt{reservoir} and \texttt{analysis} layers are under active development, and the \texttt{optimize} layer remains at an earlier stage relative to the architecture-level vision described in the repository. The paper introduces the overall OpenPRC workflow, but implementation maturity is not yet uniform across all modules.

\subsection{Current limitations}
\label{subsec:limitations}

The main limitation of the present effort is uneven module maturity. The simulation and ingestion layers are already sufficiently developed to support structured experiment setup, backend-agnostic execution, trajectory generation, and video-based feature tracking, but the downstream learning, benchmarking, and optimization layers are still evolving. The full architecture described in the repository should therefore be understood as a framework direction that is partially implemented rather than a uniformly finalized software stack.

A second limitation is that the current physics simulation path is focused on the bar-hinge model in \texttt{demlat}. The broader goal of OpenPRC is to serve as an interoperability layer for the PRC community regardless of the underlying simulator. Integrating established physics engines --- including PyBullet for rigid-body and contact-rich dynamics, PyElastica \cite{Naughton2021,Gazzola2018} for Cosserat rod assemblies, and MERLIN \cite{Liu2016,Liu2017} for quasi-static origami mechanics --- is an explicit near-term priority. Any simulator that can write trajectories to the standardized \texttt{simulation.h5} schema immediately gains access to the full downstream \texttt{reservoir}, \texttt{analysis}, and \texttt{optimize} pipeline without further modification.

\subsection{Near-term development priorities}
\label{subsec:future}

The most immediate next steps follow directly from the repository roadmap. First, the \texttt{reservoir} and \texttt{analysis} layers need to be expanded into a more complete and stable interface for task definition, readout management, benchmarking, and performance diagnostics. Second, the \texttt{optimize} layer needs to mature into a full topology and parameter search interface, closing the evolutionary design loop back to \texttt{demlat} and eventually to external simulators. Third, the shared HDF5 schema and experiment workflow should be strengthened so that simulation outputs, experimentally acquired trajectories, and data from third-party simulators can all move through the same downstream pipeline with minimal manual conversion.

Beyond module completeness, a central goal is to establish OpenPRC as a standardizing force in the PRC community --- providing a common language for describing substrates, evaluation protocols, and optimization objectives so that results from different research groups remain reproducible, scalable, and mutually comparable.

\subsection{Broader significance}
\label{subsec:impact}

Establishing a rigorous theory for physical reservoir computing requires a deep synthesis of nonlinear mechanics and machine learning, but the computational overhead --- spanning high-fidelity simulation, feature extraction, and high-dimensional optimization --- demands software engineering that often exceeds the traditional scope of mechanics research. OpenPRC is designed to lower this barrier. By organizing physics simulation, experimental ingestion, readout training, benchmarking, and optimization around a shared schema and modular workflow, it aims to make PRC studies easier to reproduce, extend, and compare across substrates and data sources. The GPU-accelerated hybrid solver enables iterative topology search and massive-batch hyperparameter tuning at speeds that were previously computationally prohibitive, allowing researchers to evolve physical designs at considerably higher efficiency than CPU-based or MATLAB-based workflows. For a field that currently relies heavily on custom scripts and study-specific pipelines, a well-structured open framework --- even one that is partially complete --- can play an important role in accelerating community-wide progress.

\section{Conclusion}
\label{sec:conclusion}

We have presented \textbf{OpenPRC}, a modular, GPU-accelerated open-source framework that unifies mechanical simulation, experimental trajectory ingestion, reservoir computing evaluation, capacity characterization, and physics-aware optimization in a single composable pipeline. The five modules --- \texttt{demlat}, \texttt{openprc.vision}, \texttt{reservoir}, \texttt{analysis}, and \texttt{optimize} --- are connected by a universal HDF5 schema that enforces reproducibility and interoperability across simulated and experimentally acquired data. The hybrid RK4--PBD solver in \texttt{demlat} achieves substantial speedup over CPU and MATLAB baselines, enabling iterative optimization and massive-batch hyperparameter exploration that were previously computationally prohibitive. The \texttt{openprc.vision} module bridges the gap between laboratory recordings and the computational pipeline, making the simulation-to-experiment interoperability claim concrete rather than aspirational.

The longer-term vision for OpenPRC extends beyond \texttt{demlat} to encompass a standardized interface for external physics engines including PyBullet, PyElastica, and MERLIN, so that any simulator capable of writing to the shared schema gains immediate access to the full downstream analysis and optimization stack. The broader goal is to provide the PRC community with the highly efficient, reproducible, and extensible software infrastructure needed to systematically relate physical design choices to computational performance --- and in doing so, to accelerate the development of a rigorous, substrate-agnostic theory of physical reservoir computing.

\section*{Availability}
OpenPRC is openly available at \url{github.com/DARE-Lab-VT/OpenPRC-dev}.
Contributions and issues are welcome via GitHub.

\section*{Acknowledgements}
The authors acknowledge the gracious support from the National Science Foundation (CMMI-2328522; EFMA-2422340).

\bibliographystyle{unsrtnat}
\bibliography{references}

\end{document}